# Integrating Model Construction and Evaluation

**Robert P. Goldman**
Department of Computer Science
Tulane University
New Orleans, LA 70118-5698
rpg@cs.tulane.edu

**John S. Breese**
Rockwell International Science Center
444 High Street
Palo Alto, CA 94301
breese@rpal.rockwell.com

## Abstract

To date, most probabilistic reasoning systems have relied on a fixed belief network constructed at design time. The network is used by an application program as a representation of (in)dependencies in the domain. Probabilistic inference algorithms operate over the network to answer queries. Recognizing the inflexibility of fixed models has led researchers to develop automated network construction procedures that use an expressive knowledge base to generate a network that can answer a query. Although more flexible than fixed model approaches, these construction procedures separate construction and evaluation into distinct phases. In this paper we develop an approach to combining incremental construction and evaluation of a partial probability model. The combined method holds promise for improved methods for control of model construction based on a trade-off between fidelity of results and cost of construction.

## 1 Introduction

Most applications of belief networks for probabilistic reasoning systems have relied on a fixed belief network. The network is constructed by the system designer (possibly in concert with a domain expert) and then used by the application to evaluate the probability of various hypotheses given observations. Recent work, much of it reported at this conference, has made clear that evaluation of such predefined, static models is not sufficient in many applications [1, 7, 6, 12]. Some drawbacks of such static models are inflexibility, lack of expressive power, and an inability to model *a priori* all possible situations [16]. One approach, which has been gaining in popularity, has been to mate a declarative model-construction component with a system for model evaluation. We refer to this approach as Knowledge-based model construction (KBMC).

In most previous KBMC systems, there has been a separation between the construction and evaluation components. The system generates a network from an expressive knowledge base which is then passed to an evaluative method. In the work described here, we present an algorithm which integrates these two components. The method described here uses a database, which describes a class of probabilistic models, to answer queries of the form: "What is the probability of proposition $x$, given evidence $y$?" Rather than building a model and then evaluating it, our approach searches through the knowledge base of model information to answer the query more directly in a deductive style of reasoning. The basic approach combines elements of query-based probabilistic inference algorithms (such as [14, 3]) with existing model construction approaches [1]. The result is an approximation algorithm for probabilistic inference, based on evaluation of a partial model at each stage of the model construction process.

The primary motivation for combining model construction with evaluation is to provide better mechanisms for control of model construction. From a decision-theoretic perspective, one wishes to continue to elaborate a model only if the benefits (quality of the answer to the query) exceed the costs (in terms of computational effort). By combining evaluation with construction, we can build an "anytime" algorithm by calculating the implications of the partially constructed model. We can then stop at any time and return partial information about the probability of a proposition. This will allow us, in turn, to take a decision-theoretic look at the control of model-building, in a simpler way than if we have to allow our model construction component to run to completion, and then control the model evaluation.

Automated model construction techniques are most appropriate where it is not practical to construct a fixed model in advance due to changes in the nature of queries or dependencies from case to case. Separating construction and evaluation is useful in a situation where we need to configure the belief network to answer a class of queries over some period. A combined construction and evaluation technique is use-ful in time-pressured, knowledge-rich domains, where time constraints make it impossible to use a large extensive model.

The paper is organized as follows. In the next section, we review the model description language our program uses. Then we present and discuss the algorithm. We discuss a sample use of the algorithm. We comment on a prototype implementation. We conclude with some discussion of research directions that this algorithm opens up.

## 2  Review of ALTERID Language

We have adopted the language of ALTERID for our approach. The basic structures are described briefly here (see [1] for a more detailed discussion). We will illustrate the constructs using relationships from a network originally presented in [2].

Deterministic relationships in the domain are represented with a set of logical formulae. A formula is *atomic* if it is of the form $P(x_1, x_2, \ldots, x_n)$ where $P$ is a relational constant and the $x_i$ are variables (lowercase) or object constants (uppercase). Facts $(P \leftarrow)$ and rules $(P \leftarrow Q)$ are defined in the normal manner for Horn-clause logic programs.

To capture the notion of mutually exclusive, collectively exhaustive, sets of outcomes for a variable, we introduce the notion of alternative outcomes. The notation $P(x, \{A, B\})$ means that for all values of $x$, exactly one of $P(x, A)$ and $P(x, B)$ is true. We will denote this set of outcomes by $\Omega$, for example, $\Omega_{P(x, \{A, B\})} = \{P(x, A), P(x, B)\}$. One of these outcomes will be indicated by $\omega_P$. For our test domain, the set of alternative outcomes is as follows:

Cancer({YES, NO},y)
Serum-Calcium({BAD,GOOD},y)
Tumor({YES, NO},y)
Coma({YES, NO},y)
Headache({YES, NO},y)

A **probabilistic dependency** is an expression of the form

$$P|_p Q_1 \wedge Q_2 \wedge \ldots \wedge Q_n = \Pr(\omega_P | \omega_{Q_1 \wedge Q_2 \wedge \ldots \wedge Q_n})$$

where $P$ is an alternative outcome expression and each $Q_i$ is an atomic formula (possibly an alternative outcome expression) and Pr is a conditional probability distribution over the alternative outcomes of $P$ given the alternative outcomes for $Q_1 \wedge Q_2 \wedge \ldots \wedge Q_n$. The dependency describes the uncertainty regarding $P$ in the state of information where $Q_1 \wedge Q_2 \wedge \ldots \wedge Q_n$ is true. For the cancer domain, we describe the conditional probability of coma given its predecessors as:

Coma({YES, NO}, y)|$_p$Tumor({YES, NO}, y)
$\wedge$Serum-Calcium({BAD, GOOD}, y)
$= \Pr(\omega_{Coma} | \omega_{Tumor}, \omega_{Calc}) =$

| COMA | YES | NO |
|---|---|---|
| $Tumor(YES) \wedge Calcium(BAD)$ | .80 | .20 |
| $Tumor(YES) \wedge Calcium(GOOD)$ | .80 | .20 |
| $Tumor(NO) \wedge Calcium(BAD)$ | .80 | .20 |
| $Tumor(NO) \wedge Calcium(GOOD)$ | .05 | .95 |

Other conditional probability relationships are represented in a similar manner. In Section 4 we illustrate the construction procedure on this example.

## 3  Algorithm MCE

The MCE (Model Construction/Evaluation) algorithm constructs and evaluates a model for a conditional probability query of the form $P(H|E)$, where $H$ is a alternative outcome statement, and $E$ is an evidence set of the form $E_1 = \omega_{E_1,i}, E_2 = \omega_{E_2,j} \ldots E_n = \omega_{E_n,k}$. All evidence relevant to the hypothesis $H$ is included in the set $E$. The output of the algorithm is a matrix describing the probability distribution for random variable $H$, given evidence in $E$ and probability model information given in the database. In addition, at any time during the operation of the algorithm, a search state can be queried to generate an approximation to the query. This bound is based on evaluation of partial probabilistic model.

MCE is an agenda-based search algorithm. In the section below we describe the search states, and the operators which can be used to yield successors of a search state.

### 3.1  Search states

Search states will contain information about the goal of the search, information about the unifications that have been done in the search, a graph which represents an expression, possibly partially-evaluated, for the target probability distribution, as it is known so far, and some control information. Formally, we describe a search state as a tuple, $S = (P^*, \Theta, G, M^*)$. We address each of the components of the search state in turn:

1. Sub-goals, $P^*$. These are formulae, which may represent random variables or categorical facts to be retrieved from the database.

   They may have associated out-edges when added by the algorithm below. This is because we add a node to the query graph (see below) only when we have found all of its parents (causal influences).

2. A substitution (most general unifier), $\Theta$. Since we will be retrieving modeling information from a deductive database, information about the binding of logical variables must be maintained.

3. A graph which represents the current form of the expression for the queried probability, $G = (V, E)$. Associated with every vertex $v \in V$ is a distribution. Note that a vertex $v$ may correspond to a set of random variables whose distributions have been multiplied together. There is an index function from formulae/random variables to graph nodes.

4. A set of formulae whose probability has been marginalized out of the above expression, $M^*$. This information is used to detect when a random variable has been prematurely marginalized out of the conditional probability expression.

The algorithm is invoked initially with a state of $(\{H, E\}, \emptyset, (\emptyset, \emptyset), \emptyset)$. The goal is construction of the graph $G$ that can be used to correctly answer the query.

## 3.2    Search actions

Search actions fall into two broad classes: those that serve to *construct* the current model and those that partially *evaluate* the model. Broadly speaking, the model construction/extension search actions take a sub-goal (a random variable), and add a corresponding node to the graph. In the process, new sub-goals may be generated, since causal influences on the current sub-goal must be found. The model construction actions are all based on the "Causal Belief Net Algorithm," in [1].

The alternative to expanding the model is to partially evaluate the probability expression. The two actions used to evaluate the graph are 1) combination of nodes, and 2) marginalizing out a random variable. The former corresponds to clustering [9], and the latter to node absorption [13]. Our treatment of the evaluation actions follows conventions introduced in the Symbolic Probabilistic Inference algorithm [14]. Marginalization is necessary to find a numerical answer to the query. Marginalizing early also may reduce the cost of evaluating the expression, by eliminating some multiplications. However, it also has the potential of wasting effort, if the marginalization is done too soon (if a node is marginalized out before all direct influences on it are found).

We now describe each search action, starting with the model construction operators, and then the evaluation operators.

### 3.2.1    Model Construction Operators

**find-prob-dependency**$(P, S)$ For $P$ a sub-goal of the search state $S$ and for each probabilistic dependency statement of the form $(A|B)$ with $B = (Q_1 \ldots Q_n)$ for which there is a substitution, $\Theta'$ extending $\Theta$ such that

$$A\Theta' = P\Theta'$$

create a new search state as follows:

$$((P^* - \{P\} \cup B\Theta'), \Theta', G', M^*)$$

$G'$ is formed by adding $P$ to $G$. A node for $P$ is added to $V$ and all edges from $P$ to nodes it causally influences are added to $E$.

Note that the search must be controlled so that **find-prob-dependency** is never applied to a sub-goal $P$ in $S$ to which the **find-in-graph** action may be applied (see below).

**prove-goal**$(P, S)$ $P$ an atomic formula. For every $\Theta' \in \mathbf{prove}(P, \Theta)$, create a new search state as follows:

$$S' = (P^* - \{P\}, \Theta', G, M^*)$$

In the new search state the subgoal $P$ has been removed because it has been proven.

**prove**$(F, \Theta)$ is a standard Prolog-style horn-clause deduction system. It returns a set of substitutions every one of whose elements is some $\Theta'$ such that $F\Theta'$ follows from the contents of the database, and $\Theta'$ is an extension of the previous substitution, $\Theta$.

**find-in-graph**$(P, S)$ there is a node $N \in V$, and a substitution $\Theta'$, extension of $\Theta$ such that

$$N\Theta' = P\Theta'$$

create a new search state as follows:

$$S' = (P^* - \{P\}, \Theta', G', M^*)$$

$G'$ differs from $G$ only in the addition of out-edges from $N$ to children of $P$.

In this case we have found a new path from some child node to a random variable that has already been included in the model.

**detect-marg-error**$(P, S)$ there is a node $N \in M^*$, and a substitution $\Theta'$, extension of $\Theta$ such that

$$N\Theta' = P\Theta'$$

then **Fail**. This search state is invalid because some node has been marginalized out before all of its children have been included in the model.

### 3.2.2    Evaluation Operators

**multiply**$(N, N', S)$ For $N, N' \in V$. The **multiply** action merges together two graph nodes, and in parallel, multiplies together their distributions to give a new distribution (possibly with a larger state space).

The result of the multiply action is a new search state as follows:

$$S' = (P^*, \Theta, G', M^*)$$

It is exactly as the previous search state, but with an altered $G$. We replace $N$ and $N'$ with a new node $N^{N \cdot N'}$. We replace all edges $(x, N), (x, N'), (N, x), (N', x)$ with new edges $(x, N^{N \cdot N'})$ and $(N^{N \cdot N'}, x)$. We multiply the probability distributions of $N$ and $N'$ to give the matrix for $N^{N \cdot N'}$, whose dimension is the union of the dimensions of $N$ and $N'$.

**margin**$(F, S)$ For $F$ a formula whose state is referred to in exactly one $N \in V$. Marginalize out the random variable $F$ to give a new search state as follows:

$$S' = (P^*, \Theta, G', M^* \cup \{F\})$$

The new graph, $G'$, is the same as $G$, but the values of random variable $F$ have been marginalized out of the node which is indexed under $F$. We mark the state to indicate that $F$ has been marginalized out. This is used to detect errors in action **detect-marg-error**.

## 3.3 Evaluating Partial Models

The benefits from a combined constructor/evaluator arise from the ability to monitor the progress of the construction algorithm in terms of its progress towards answering the query. For example, in time pressured domains or with extremely large knowledge bases, it may be necessary to cease model construction activity before all causal and diagnostic links have been explored. We need to be able to access each search state and use the currently constructed model to provide partial information (e.g. probability bounds) about the query.

In the model construction algorithm described here, each search state contains a network $G_S = (V, E)$ representing a partial model, that if successfully completed, will be capable of answering the query exactly. Let $G_f = (V_f, E_f)$ be a complete and consistent network constructed through a successful termination of the MCE algorithm. Let $\mathcal{G}_S$ be the set of all such possible successful completions:

$$\mathcal{G}_S = \{G_f | G_f \in Descendants(S)\}$$

where $Descendants(S)$ are the search states accessible from state $S$. Evaluation of the partial model $G_S$ consists of making probability statements about distributions consistent with all networks in $\mathcal{G}_S$. Obviously, we do not have the completion set $\mathcal{G}_S$ to work with when we do the partial evaluation, but we can make some statements about its characteristics based on the the current state of the search alogorithm and the query. In general, this involves making some assumptions about how the model construction sequence will terminate.

We have identified three modes for evaluation of partial models during a construction sequence:

### 3.3.1 Correct Scoring

In this alternative, we make no conclusions regarding the ultimate distributions without a conclusive proof of correctness. Thus, probability statements must be consistent with all possible completions. Unfortunately, for most construction algorithms, including ours, a highly evocative diagnostic link may be added to a model at any time rendering proper bounds noninformative.

### 3.3.2 Default Scoring

Here we assume that any completely specified fragment of the network currently being constructed will constitute the finally constructed network. We look for all nodes whose immediate predecessors and indirect predecessors are fully specified in $G_S$. Any nodes in $G_S$ that rely on subgoals still in $P^*$ are not included. We apply an exact algorithm to this subnet. This procedure is equivalent to assuming that all remaining subgoals in $P^*$ will fail. Another way of viewing this assumption is that at the time we ask for the answer, the currently constructed model is all that is available and should therefore be used. Obviously, as new subgoals are proven the structure and results of the query will change. This nonmonotonic behavior of the partial evaluation reflects the same concerns identified in previous defeasible probabilistic reasoning schemes [8, 10].

### 3.3.3 Interval Scoring

In this method, we treat the partially constructed model as an interval-based network. In this method, we treat nodes that have been identified but whose parameters are as yet unspecified as having probabilities in $[0, 1]$. We process the resulting intervals on the query using node absorption and arc reversal procedures developed for interval probabilities [4, 5]. In contrast to default scoring, we assume that the subgoals in $P^*$ will succeed, but the complete specification of the parameters of the model is incomplete. This techniques also exhibits nonmonotonic behavior.

We are currently experimenting with the behavior of these alternatives as mechanisms for partial evaluation. We are continuing to further refine partial evaluation methods applicable to particular types of search.

## 3.4 Search Control

Two issues dominate the control of search for MCE. The first is management of the search for a model. As soon as possible, we would like to discard partial models which do not fit the query. We argue that this issue is similar to the issue of discarding inappropriate proof trees in automated deduction, and will have little to say about this here.

The second search issue is the scheduling of **margin** actions relative to other search actions. We are indebted to Schachter, et. al. [14] for this perspective on model evaluation. Two search decisions must be made in choosing to employ the **margin** operation. The first is when the operation is likely to be correctly applicable. The search algorithm should have heuristic information which prevents it from prematurely marginalizing out nodes. Assuming that one avoids incorrectly removing nodes,the second decision addresses selecting the order of marginalization to minimize the cost of evaluating the query expression.

We are just beginning to explore the issues of search control for the MCE algorithm. We return to the subject of search control in the section on future work.

## 4    Example

To give a flavor for the use of the algorithm, we will describe one way of answering a query from the simple medical diagnosis domain introduced in Section 2.[1] We will assume that we have observed that our patient, Sam, has a headache and is in a coma. We would like to assess the probability that Sam has cancer.

We create a search state with the following sub-goals:

{(cancer ?alt SAM), (headache HEADACHE SAM), (coma COMA SAM)}

The search state also has an empty digraph ($G = (\emptyset, \emptyset)$), empty substitution ($\Theta = \emptyset$) and empty list of marginalized nodes ($M^* = \emptyset$).

Note that we have felt free to direct the search in this example by hand, for instructional purposes, and also to keep the number and size of search states manageable.

**First search action**    Let us assume that the algorithm chooses the first of these subgoals, the query to be investigated, and the **find-prob-dependency** search action.[2]

In the database, there is a probabilistic dependency statement which gives a prior probability for the query (cancer ?alt sam). This yields a new search state. In this new search state, the only two remaining subgoals correspond to the two pieces of evidence. The node contains a digraph which contains only one node which represents the query variable. There are a number of bindings in the substitution. Because the logic-programming aspects of this example are simple, and in the interests of brevity, we will not further discuss the management of substitutions.

We also now have a partial probability model. There is a single node for cancer in the network. Partial evaluation of this model at this point can proceed in several ways as discussed above. Under the *default* method the incremental answer is just the prior. We assume that additional subgoals will fail in the sense they will add no relevant dependencies to the model.

**Second search action**    At this point, with only one node in the graph (and that the query node), there are no nodes available for multiplication or marginal-ization. So the search algorithm will apply the **find-prob-dependency** action again, to find a probabilistic dependency for another sub-goal. This time we search for causal influences on the headache observation.

We retrieve from the database the statement which reports that headache depends on the presence or absence of a tumor. The conditional probabilities of **headache** and **no-headache** based on the possible values of **tumor** are also retrieved from the database.

The resulting search state has three sub-goals: a new goal for the predecessor of the headache variable, **tumor**, and the previously-existing one for the remaining piece of evidence from the initial query: (COMA **COMA SAM**). The two nodes in the digraph now are: one for **headache** and one for **cancer**, whose connection is still unknown.

Partial evaluation at this point still returns the prior. There is no dependency yet uncovered by the algorithm.

**Third and fourth search actions**    There are still no nodes available for evaluation actions, so we choose to apply **find-prob-dependency** to the **tumor** subgoal. We find that **tumor** depends on the outcome of **cancer**. Cancer is once again added to the list of subgoals. Note that the connection between **cancer** and **tumor** nodes has not yet been found.

Applying the **find-in-graph** action to the new **cancer** subgoal uncovers the connection. The node for the **cancer** variable (which we added in our first search action) is found already in the digraph. The resulting search state has only one remaining sub-goal: **coma**. The corresponding digraph is given as Figure 1.

**Early evaluation (actions 5,6)**    We are now presented with the first opportunity to employ the evaluation actions. We may be certain by inspecting the database that the **headache** node will not causally influence any other nodes.[3] Accordingly, we have the opportunity to multiply it into its parent, and marginalize it out of the graph. Note that since we have observed the value of the **headache** node, the effect of marginalizing it out is only to remove a number of zeros from the combined node's matrix. The result of these two actions may be seen in Figure 2.

At this point the partial evaluation action could also be undertaken under the *default* method, assuming that network illustrated in Figure 2 will be the final network. Calculation of the query probability would proceed by calculating the joint probability of **cancer** and **tumor** and summing out **tumor**.

**Completing the diamond (actions 7-10)**    At this point, we could either multiply the **cancer** node into

---

[1]There will, in general, be many sequences of actions which would produce an answer to the query.

[2]Note that it is necessary always to make sure that the find-in-graph action would fail before using find-prob-dependency. This is easy to achieve and we will let this pass without comment from now on.

[3]In Section 3 we discuss ways to use the structure of the database to control search.

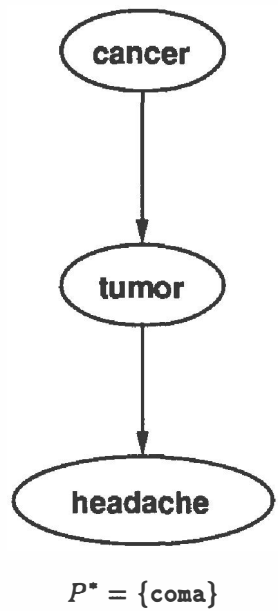

$P^* = \{\text{coma}\}$

Figure 1: The search state after four search ops.

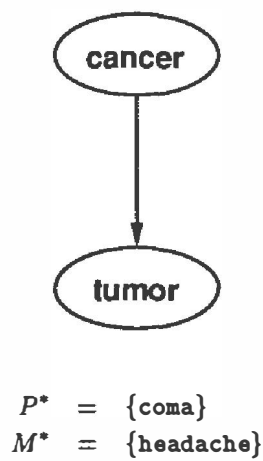

$P^* = \{\text{coma}\}$
$M^* = \{\text{headache}\}$

Figure 2: The search state after two evaluation operations.

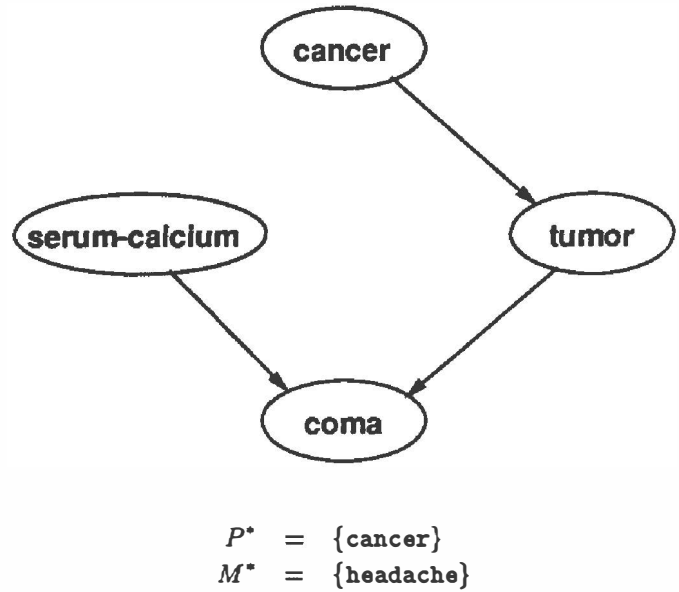

$P^* = \{\text{cancer}\}$
$M^* = \{\text{headache}\}$

Figure 3: The search state just before the diamond is completed.

the node for tumor, or we could apply the **find-prob-dependency** action to the remaining sub-goal, coma. Assuming we choose the latter, we will find that the coma variable depends on tumor and on serum-calcium.

Further applying the **find-in-graph** action to the tumor sub-goal and the **find-prob-dependency** action to the serum-calcium sub-goal, we arrive at the search state depicted in Figure 3.

At this point we have an incomplete network. We know serum-calcium is in the network, but we do not know its probabilities. We apply a combination of interval and exact transformations to the diagram to calculate a bound on the query probability, getting a result that the probability lies in the interval [.20, 3744]. Note that this result ignores the possibility of a dependency between cancer and serum-calcium.

A final **find-in-graph** action will complete the diamond.

**Completing the query** The process of answering the query may now be completed by a series of evaluation actions. We suggest the following series, but others are also suitable:

1. marginalize out serum-calcium. This leaves a generalized distribution at coma which depends only on cancer and tumor.

2. marginalize out tumor

3. multiply cancer into coma

4. marginalize out coma and normalize to get the distribution for cancer, $P(\text{cancer}|Ev) \approx .438$.

## 5   Implementation

The algorithm described here has been implemented in (Sun/Lucid) Common Lisp, running on Sun SPARC-stations. It has been tested on the example given here and other examples like it. The program has been written in several different modules: one that manages the deductive database; one that manages the matrix operations; and one which manages the search operations. Influence diagram processing is performed with IDEAL [15]. We thank Peter Norvig for allowing us to use his Prolog interpreter in Common Lisp for our deductive retrieval [11].

The code is still in prototype version, and many opportunities for optimization remain. The bounding calculus has not been integrated into the construction cycle. We are still using only a generic agenda-based search algorithm. For this algorithm to be practically usable, we will have to extend the code for agenda maintenance to better control the search. Elsewhere in this paper we have suggested search control methods we believe will be successful for this program. We will be investigating these heuristics and their interaction with other aspects of the work (.e.g interval processing). We will also be developing better implementations of existing elements of the system.

## 6   Future Directions

The work described in this paper is continuing on a number of avenues. We will be conducting experimental tests to explore the behavior of the algorithm over several databases. In particular, we wish to explore the interaction of partial evaluation with various methods for search control. A related issue revolves around maintaining probability interval information in product form for generalized distributions.

To avoid premature marginalization, we are experimenting with a technique which makes use of information about the structure of the knowledge base. We suggest an application of the technique of marker passing, treating the rule base as a graph. There will be nodes corresponding to alternative outcome statements. There would be edges from alternative outcome statements to causal influences, corresponding to the probabilistic dependency statements. As is characteristic of marker-passing, the values of variables would be ignored — edges would be drawn everywhere there was a possible probabilistic dependency relation to occur. Before carrying out the search for a particular query, nodes corresponding to query variables would be marked. Marks would be propagated from children to possible parents. Marks would have limited "memory" to cut off cycles. Each type of node would have a counter. Nodes would not be marginalized out until a number of children equal to the number of marks had been found (or until all possible CBN operations were done). This could be an over-cautious heuristic (especially in the case of rule-bases with much recursion), but should prevent premature marginalization. This technique can, of course, be 'outwitted,' by poorly-structured databases (ones where there are few predicates but many propositions), but well-known techniques for improving Prolog programs will also make this heuristic more accurate.

We would like to complement a technique like that discussed above with a search control method which would weigh the chance of premature marginalization against its benefits (reduction in the dimensionality of the matrices, and hence the number of multiplications). As mentioned in the previous section, we are also interested in taking an "anytime" approach to the MCE algorithm, taking into account the tradeoff between further model construction and evaluation actions, and termination of search with estimated responses to a query.

The present version of the algorithm assumes that evidence relevant to the query is identified in the query. Previous work [1] conducted a search for such evidence in the database. We wish to investigate the tradeoff in search efficiency for these alternatives.

As discussed in the introduction, the ultimate goal of this research is to provide a facility for informed control of model construction. A construction procedure that can evaluate its progress toward answering a query is an important step in this direction.